\documentclass[a4paper,twoside]{article}

\usepackage{epsfig}
\usepackage{subcaption}
\usepackage{calc}
\usepackage{amssymb}
\usepackage{amstext}
\usepackage{amsmath}
\usepackage{amsthm}
\usepackage{multicol}
\usepackage{pslatex}
\usepackage{apalike}
\usepackage{ragged2e}
\usepackage{multirow}
\usepackage{algorithm2e}
\usepackage[bottom]{footmisc}
\usepackage{SCITEPRESS}     

\begin{document}

\title{Region-Transformer: Self-attention Region Based Class-agnostic Point Cloud Segmentation}


\author{\authorname{Dipesh Gyawali\sup{1}, Jian Zhang\sup{1} and Bijaya B. Karki\sup{1}}
\affiliation{\sup{1}School of Electrical Engineering and Computer Science, Louisiana State University, Baton Rouge, LA 70803, USA}
\email{\{dgyawa1, jz, bbkarki\}@lsu.edu}
}


\abstract{Point cloud segmentation, which helps us understand the environment of specific structures and objects, can be performed in class-specific and class-agnostic ways. We propose a novel region-based transformer model called Region-Transformer for performing class-agnostic point cloud segmentation. The model utilizes a region-growth approach and self-attention mechanism to iteratively expand or contract a region by adding or removing points. It is trained on simulated point clouds with instance labels only, avoiding semantic labels. Attention-based networks have succeeded in many previous methods of performing point cloud segmentation. However, a region-growth approach with attention-based networks has yet to be used to explore its performance gain. To our knowledge, we are the first to use a self-attention mechanism in a region-growth approach. With the introduction of self-attention to region-growth that can utilize local contextual information of neighborhood points, our experiments demonstrate that the Region-Transformer model outperforms previous class-agnostic and class-specific methods on indoor datasets regarding clustering metrics. The model generalizes well to large-scale scenes. Key advantages include capturing long-range dependencies through self-attention, avoiding the need for semantic labels during training, and applicability to a variable number of objects. The Region-Transformer model represents a promising approach for flexible point cloud segmentation with applications in robotics, digital twinning, and autonomous vehicles.}

\onecolumn \maketitle \normalsize \setcounter{footnote}{0} \vfill

\section{\uppercase{Introduction}}
\label{sec:introduction}

Point cloud segmentation is an imperative technique to understand 3D surroundings and objects, with applications in robotics \cite{robotics}, automation \cite{autonomous}, digital twinning \cite{digitaltwin}, VR/AR \cite{ar}. Most existing methods perform class-specific segmentation \cite{qi2017pointnet} \cite{pointnetplusplus} \cite{yang2019learning} \cite{pointtransformer} requiring semantic labels. However, class-agnostic segmentation without prior object knowledge is more flexible yet challenging.

Recently, self-attention networks \cite{pointtransformer} have shown promise for point cloud tasks by capturing contextual information. And region-growth approaches enable adaptive segmentation determination. However, self-attention has not been explored to enhance region-based segmentation. Our key insight is combining self-attention with region-growth can improve class-agnostic point cloud segmentation.

We propose a Region-Transformer model, which utilizes self-attention in local neighborhoods to iteratively expand/contract segments by adding/removing points likely belonging to the same instance. This model provides two key advantages over previous methods: 1) Attention on local regions captures finer relationships versus global context, and 2) Region growth allows flexible segmentation boundaries using neighborhood information. Our experimental studies show that Region-Transformer significantly outperforms previous class-specific and class-agnostic methods and thus demonstrate the benefits of our proposed approach.

In this work, our main contribution includes the following.
\begin{itemize}
\item  We leverage the power of the self-attention mechanism combined with the region-growing approach to segment an environment ranging from small-scale to large-scale data completely.
\item We don't need semantic labels to train the model that provides flexibility in segmenting any number of objects in an environment.
\item We capture local contextual information for each point inside a region that helps identify long-range point cloud data dependencies.
\end{itemize}


\section{\uppercase{Related Works}}
\label{sec:literaturereview}
Research in point cloud segmentation has primarily focused on semantic and instance segmentation. Semantic segmentation classifies each point within 3D data, while instance segmentation assigns points to a specific instance without class labels. Class-specific segmentation has been more extensively studied than challenging class-agnostic segmentation in varied real-world environments.

Point cloud data features range from XYZ positions to geometric aspects like normals and curvatures. Techniques for analyzing these features include patch stitching and octree-based hierarchical representations \cite{Gumhold2001FeatureEF} \cite{guoyulan} \cite{5703032} \cite{zhou2021fast} \cite{s19051248}. \cite{6758588} discuss datasets and methodologies for point cloud segmentation.

Deep learning has emerged as a significant method for 3D point cloud segmentation, with approaches including projection-based, discretization-based, point-based, and proposal-based methods \cite{9687584} \cite{qi2017pointnet} \cite{guo2020deep}. \cite{guo2020deep} further explore neural networks for 3D tracking, shape classification, detection, and segmentation.

Few-shot learning, neighborhood information, and class-agnostic approaches are also being explored for segmentation \cite{zhao2021fewshot} \cite{Engelmann_2019} \cite{9813403} \cite{sharma2020classagnostic}. Unsupervised methods and region-growing approaches address segmentation without labels, focusing on features and iterative calculations \cite{Xiao_2023} \cite{isprs-archives-XLII-3-W10-153-2020} \cite{lrgnet}.

Recently, transformers have been applied to point cloud data, leveraging their success in NLP and Computer Vision \cite{gyawali2023lrtransformer} \cite{pointtransformer}. Our research combines self-attention mechanisms with region-growing approaches for class-agnostic segmentation, utilizing the transformer architecture's adaptability to varying input data. Consequently, applying self-attention operations to 3D data is a logical choice, given that point clouds are collections of points within 3D space.

\section{\uppercase{Methodology}}
\label{sec:methodology}

The methodology includes problem formulation, point transformer block, interaction of self-attention and region growth, data preparation and inference.


\subsection{Problem Definition and Formulation}

We formulate point cloud segmentation as an iterative region-growing problem using a learned neural network function \textit{f}. Given a point cloud \textbf{P} with \textit{N} points represented by \textit{F} features, the goal is to assign an instance label \textbf{L} to every point. The region-growth starts from a seed point  $p\textsubscript{seed} \in $ \textbf{P} and progressively adds points $P\textsuperscript{*}\subset $ \textbf{P} belonging to the same instance to expand the region. At each step, \textit{f} transforms the input points $C\textsubscript{k}$ to output points $C\textsubscript{k+1}$. Initially, \textit{C\textsubscript{0}} = {\textit{p\textsubscript{seed}}}, until $C \rightarrow P\textsuperscript{*}$, indicating the full instance is segmented. The point cloud has 13 features, including XYZ positions, RGB colors, normals, and curvatures. Normals and curvatures are computed using PCA \cite{asao2022curvature} on local neighborhoods. Room dimensions also normalize XYZ coordinates. In total, each point is represented by a 13D feature vector.

\begin{figure*}[!htb]
  \centering
   {\epsfig{file = 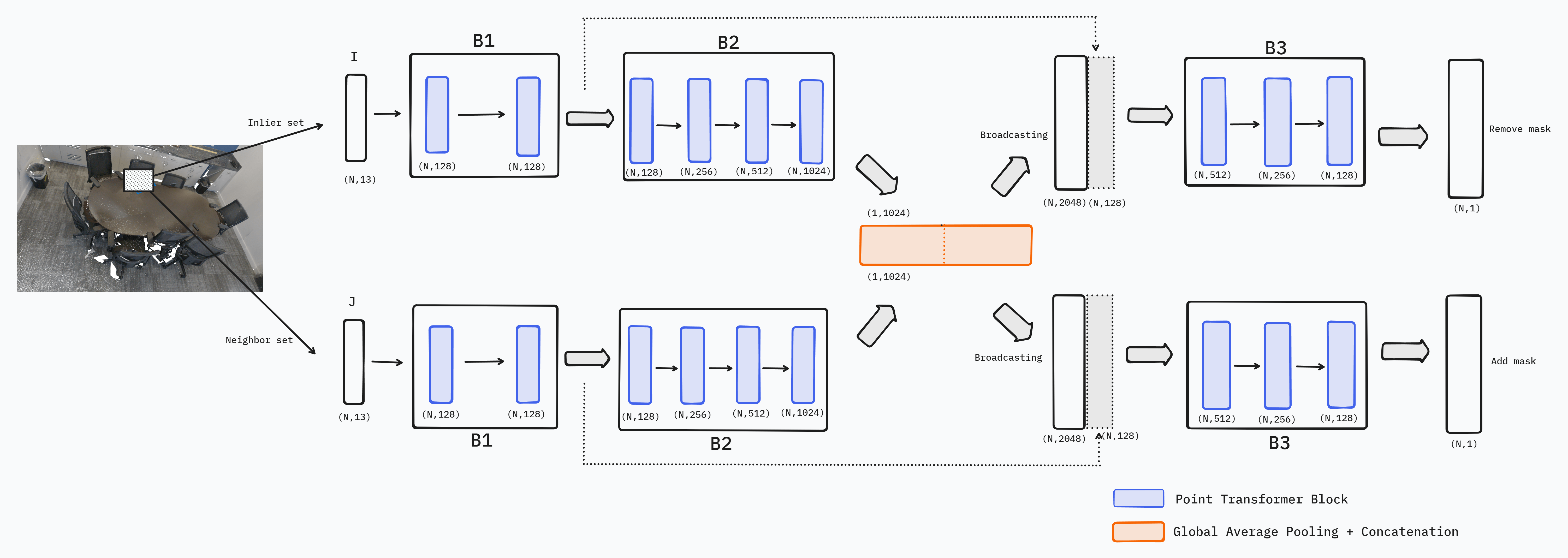, width = 15.8cm}}
  \caption{Region-Transformer network architecture for class-agnostic segmentation. Block B1 generating (128,128) features and B2 generating (128,256,512,1024) features act as encoders for inlier and neighbor sets. Block B3 generating (512,256,128) features acts as a decoder. Points from B1 and B2 are average-pooled, and inlier and neighbor sets are concatenated together to form a bottleneck. The encoded features are broadcasted into N number of points, and features from B1 output are concatenated to broadcasted features to get positional information of each point. In the last layer, the add and remove mask predictions are made.}
  \label{fig:example2}
 \end{figure*}

\subsection{Network Architecture}
The core component of the network architecture is the point transformer \cite{pointtransformer}, a neural network designed to capture both local and global contextual information of each point, considering its neighboring points. This information is crucial in determining whether neighboring points should be included or excluded in the segmentation process. As shown in Figure 1, the network consists of two branches - an inlier branch and a neighbor branch - which receive inputs of inlier and neighbor point sets. The sets pass through encoder blocks B1 and B2 to generate latent feature vectors per set. The concatenated vectors are broadcasted and decoded by B3 \cite{lrgnet}.

The Point Transformer block facilitates the exchange of information between localized feature vectors, allowing adaptation to spatial arrangements and features in 3D space. The core Point Transformer layer utilizes a self-attention mechanism to relate each point to its local neighborhood points, as shown in Figure 2. This captures contextual information to determine whether to include or exclude points during segmentation.

\begin{figure}[!h]
  \centering
   {\epsfig{file = 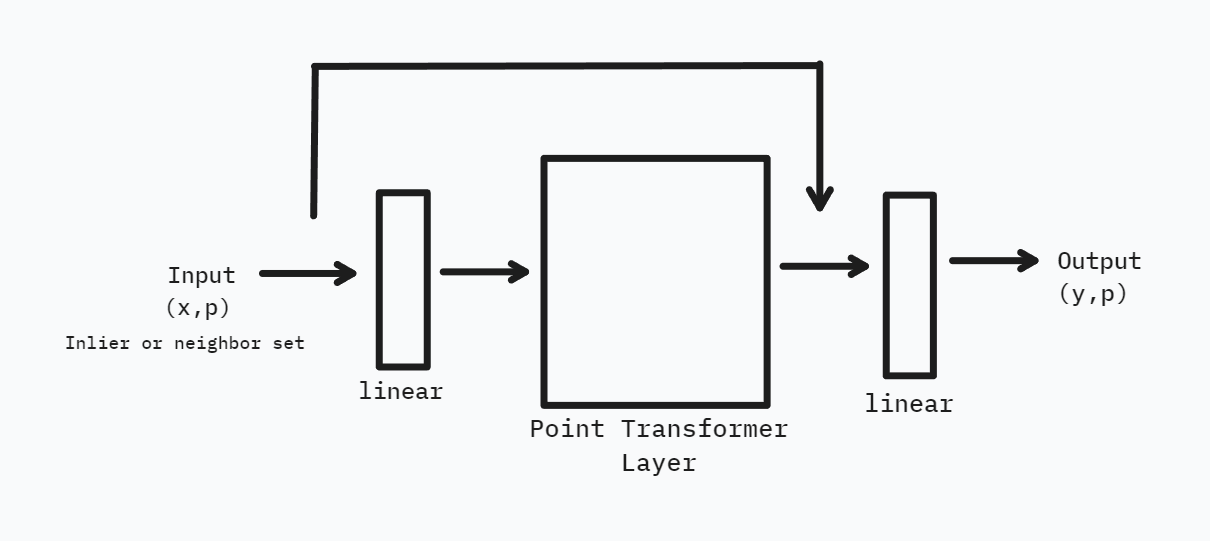, width = 7cm}}
  \caption{Point Transformer Block.}
  \label{fig:example3}
 \end{figure}

The Point Transformer layer captures each point's local and global contextual information by considering its neighboring points. For this, the layer utilizes a self-attention mechanism. Specifically, self-attention is applied locally within a predefined neighborhood (e.g., k-nearest neighbors) around each point. This allows focusing on and aggregating features from the most related subset of neighbors(16) rather than all points. The self-attention procedure uses mapper functions to transform input point features into queries, keys, and values \cite{vaswani2023attention}. Attention weights between the query and keys are calculated. These attention weights determine how much each value vector contributes to the output aggregated feature for the central point. In addition, a relative positional encoding $\delta$ is applied to retain positional information of each point in the 3D space. This encoding uses a parameterized function of the difference between point coordinates given as

\begin{equation}\label{eq1}
    \delta = \beta(p\textsubscript{i} - p\textsubscript{j})
\end{equation}

where \textit{p\textsubscript{i}} and \textit{p\textsubscript{j}} represent 3D position values for points i and j, and $\beta$ represents three fully connected layers with two nonlinear ReLU in between. The point transformer is based on self vector-attention \cite{zhao2020exploring} given as

\begin{equation}\label{eq2}
    y\textsubscript{i} = \sum_{x_j\in X(i)} \rho(\gamma(\varphi(x_i)-\psi(x_j)+\delta)) \odot (\alpha(x_j) + \delta)
\end{equation}

The self-attention procedure is defined in Equation 2, where $\varphi$, $\psi$, $\alpha$ transform input features, $\rho$ normalizes, $\delta$ encodes position, and $\gamma$ maps attention weights to aggregate features.
$X(i) \subset X$ represents the local neighborhood of point \textbf{x\textsubscript{i}}. Self-attention is applied to each point's neighborhood to focus on similar local regions rather than global contexts. The mapper $\gamma$ uses multilayer perceptrons to generate attention weights.

The output is a new feature vector for each point with selectively aggregated contextual information from its local neighborhood. These features can better determine relationships between nearby points to aid the region segmentation process. The ability of self-attention to capture dependencies based on feature similarity rather than spatial proximity helps the region grow according to semantic instance boundaries. The localization also allows finer segmentation precision.

\subsection{Self-attention in Region-Growth}

Our method includes encoder-decoder network which is trained to learn effective region-growth with a binary cross-entropy loss on the addition and removal predictions. The loss compares predicted point inclusion/exclusion probabilities with ground truth labels given as

\begin{equation}\label{eq3}
   \begin{split}
        \mathcal{L}
         = -\frac{1}{I} \sum_1^I [x_i\log{\hat{{x}\textsubscript{i}}} + (1-x_i)\log{(1-\hat{{x}\textsubscript{i}})}] \\ 
         -\frac{1}{J} \sum_1^J [y_j\log{\hat{{y}\textsubscript{j}}} + (1-y_j)\log{(1-\hat{{y}\textsubscript{j}})}]   
\end{split}
\end{equation}

Our key insight enabling improved segmentation performance is using self-attention within the context of iterative, neural network-guided region-growth. Local point neighborhoods are defined around seed points using a radius threshold. Self-attention is applied to every neighborhood, enabling each point to aggregate features from its local context. It captures nuanced geometric relationships between a point and its neighbors that standard features miss. This higher-order feature representation is input to the Point Transformer network to predict iterative growth decisions. So self-attention does the "heavy lifting" to equip the network with finer-grained neighborhood characterization for superior growth predictions. The Point Transformer network analyzes attention-enhanced local features to predict binary masks, indicating which neighborhood points should be added/removed to grow the region. Based on these iterative add/remove decisions, new local neighborhoods are extracted around the grown region's updated seed points. Self-attention and neural feature processing are repeated in the new neighborhoods, further evolving the regions to capture more points belonging to the same instance. 

The key novelty is using the neural predictions from intermediate attention-augmented features to actively determine how regions evolve rather than relying solely on hand-crafted similarity metrics. This dynamic interaction helps address limitations of both attention and region growth in isolation.

\subsection{Data Preparation and Simulation}
The S3DIS \cite{s3dis} and Scannet \cite{scannet} datasets containing point cloud labels generate simulated training data. The simulation follows a region-growth approach based on \cite{lrgnet} with PyTorch implementation. Data augmentation is applied, including random flipping, rotation, and introducing mistake probability($\theta$) noise. 844 million point clouds from 3.5 million instances are generated as S3DIS training data, in addition to 17 million validation sets. Similarly, 741 million point clouds from 5.0 million instances are generated as Scannet training sets.
The simulation grows regions starting from random seed points, iteratively aggregating nearby points sharing the same instance label. Noise is gradually reduced over the region growth iterations to promote convergence while preventing overfitting. This process creates training data mimicking realistic instance segmentations. The validation data evaluates generalization. Each instance is unique despite identical class labels. The approach synthesizes sufficiently large and diverse labeled data for effectively learning the region growth transformations.

\subsection{Inference}
The inference technique and conditions are derived from \cite{lrgnet}. During inference, segmentation is performed by iteratively adding and removing points from an initial seed region until all points are labeled. The trained transformer network outputs addition and removal predictions to grow regions. The process continues until one termination criteria:

\begin{itemize}
    \item No neighboring points are remaining to be assigned to the region.
    \item The points set to be added are empty.
    \item There is no expansion of region for two consecutive steps.
\end{itemize}

On termination, the final region is assigned an instance label and then reset with a new seed. Seeds are strategically selected as the point with the lowest curvature for consistency \cite{81fc0f1e9d33404e8279328c15c08bca}. For robustness, segments with few points (less than 8) do not form new instances. Instead, points adopt the label of nearest neighbors. This prevents the loss of points between larger segmented instances \cite{hu2019learning} \cite{xie2021unseen}. 

In summary, inference progressively segments the point cloud into instance-labeled regions by learned prediction of what points to add/remove at each iteration. Termination and seed selection strategies maximize completeness, consistency, and efficiency.

\begin{figure*}[!htb]
  \centering
   {\epsfig{file = 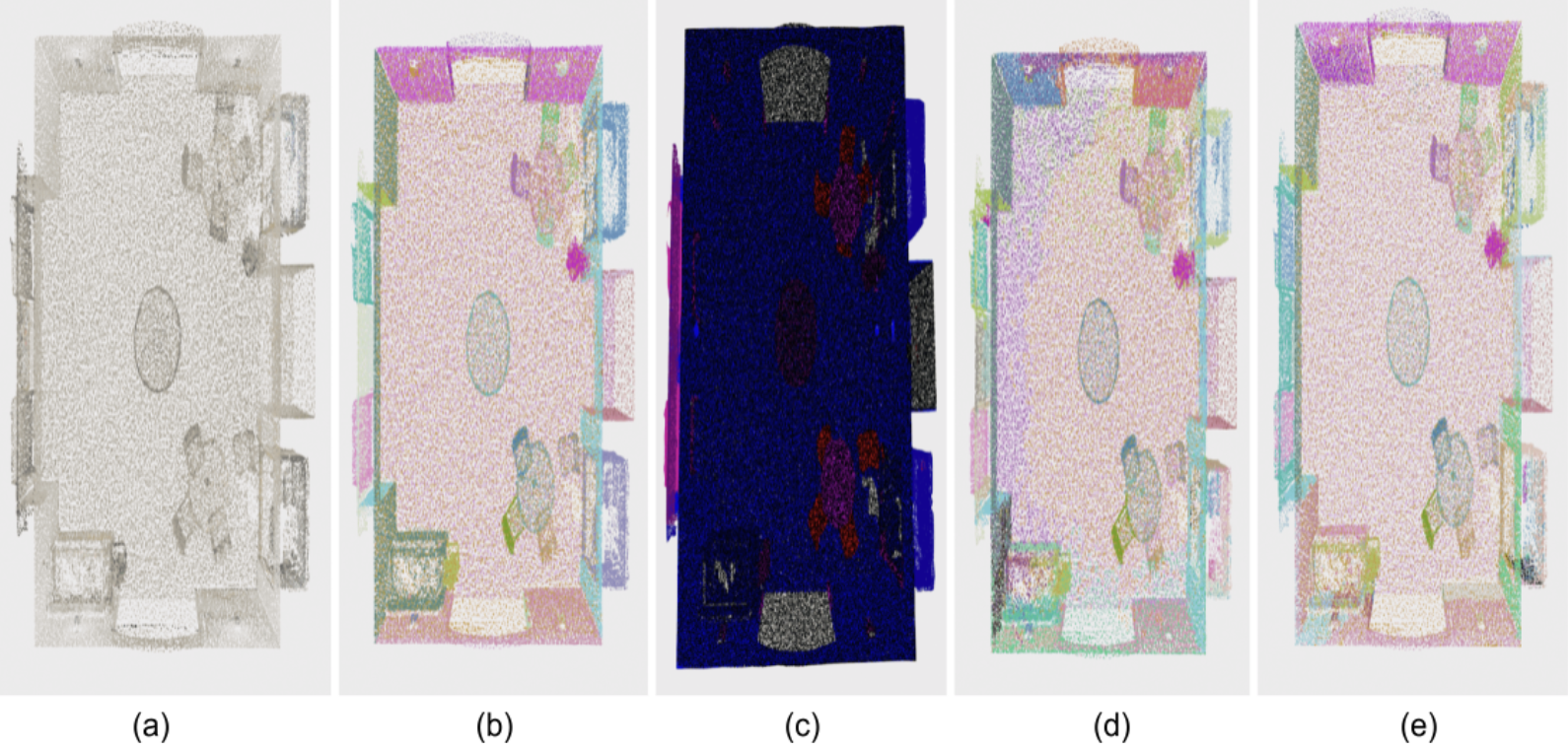, width = 14cm}}
  \caption{Object-agnostic segmentation results across Scannet a) Raw point cloud original visualization (b) Ground truth original segmentation (c) PointNet++ segmentation (d) LRGNet segmentation (e) Region-Transformer (Our Method).}
  \label{fig:example5}
 \end{figure*}

\begin{figure*}[!htb]
  
  \centering
   {\epsfig{file = 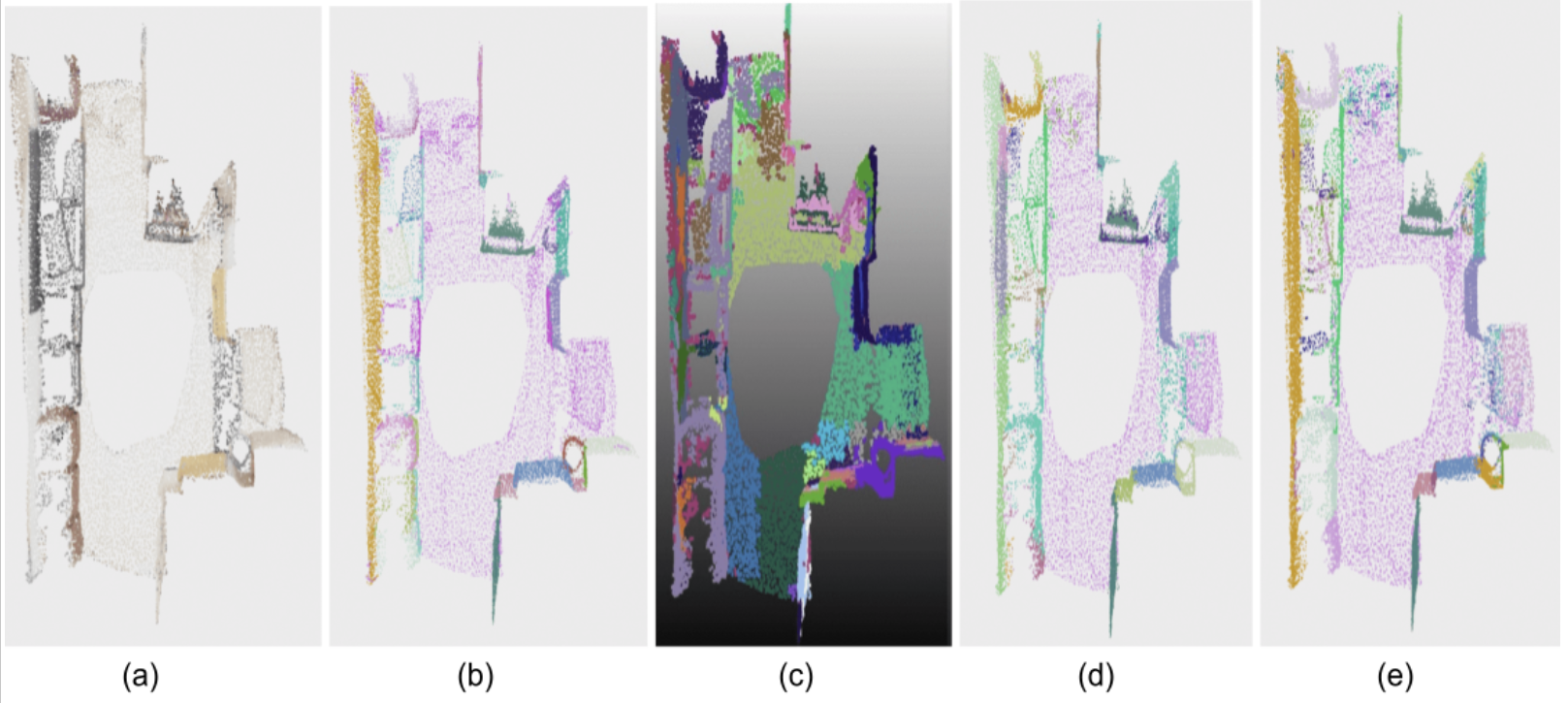, width = 14cm}}
  \caption{Object-agnostic segmentation results across S3DIS a)Raw point cloud original visualization (b) Ground-truth original segmentation(c) PointNet++ segmentation (d) LRGNet segmentation (e) Region-Transformer (Our Method).}
  \label{fig:example6}
 \end{figure*}

\begin{figure*}[!htb]
  \centering
   {\epsfig{file = 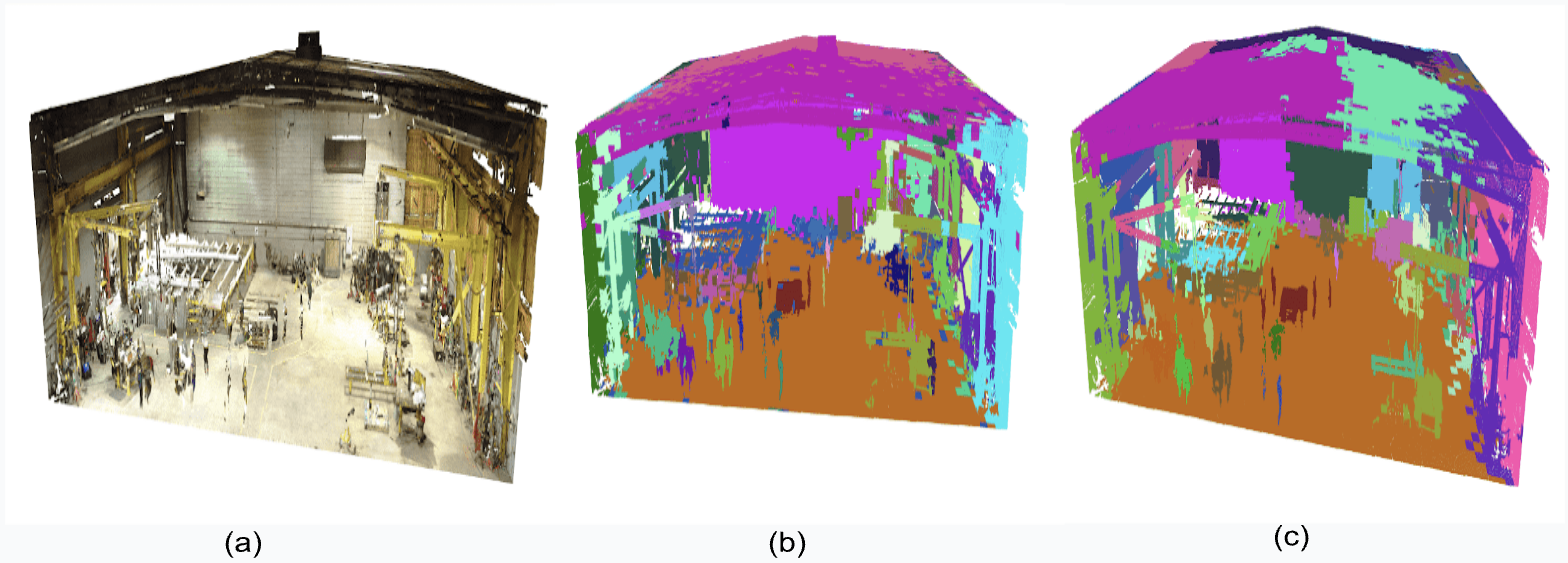, width = 14cm}}
  \caption{Object-agnostic Segmentation results on real-world, large-scale factory data a) Raw point cloud original visualization (b) LRGNet segmentation (c) Region-Transformer (Our Method).}
  \label{fig:example7}
 \end{figure*}

\section{\uppercase{Experiments and Results}}
\label{sec:experimentsandresults}

The Region-Transformer model was tested for segmentation on indoor (S3DIS, Scannet) and outdoor scenes, using clustering metrics (ARI, AMI, NMI) and general metrics (mean IOU, Precision, Recall). Comparisons were made with class-agnostic and class-specific segmentation methods. Implemented in Pytorch, the model used Adam Optimizer, which was trained over 90 epochs on NVIDIA RTX A6000 GPU, and the training took about seven days.

\begin{table*}[ht]
\caption{Comparison between models using Scannet as training and S3DIS as test.}\label{tab:example1} \centering 
\begin{tabular}{l c c c c c c c} 
\hline\hline 
\textbf{Type} & \textbf{Method} & \textbf{ARI} & \textbf{AMI} & \textbf{NMI} & \textbf{Precision} & \textbf{Recall} & \textbf{mIoU} \\ [0.5ex] 
\hline 
\multirow{5}{*}{\rotatebox[origin=c]{90}{\parbox{25mm}{\centering class-\\dependent}}}
 & PointNet \cite{qi2017pointnet} & 0.38 & 0.48 & 0.58 & 0.18 & 0.17 & 0.25 \\
 & PointNet++ \cite{pointnetplusplus} & 0.40 & 0.56 & 0.62 & 0.15 & 0.22 & 0.31 \\
 & 3D-BoNet \cite{yang2019learning} & 0.68 & 0.72 & 0.75 & 0.20 & 0.29 & 0.35\\
 & JSIS3D \cite{pham2019jsis3d} & 0.63 & 0.73 & 0.74 & 0.28 & 0.29 & 0.36 \\
 & Point Transformer \cite{pointtransformer} & 0.69 & 0.73 & 0.75 & 0.44 & 0.44 & 0.47 \\
\hline
\multirow{5}{*}{\rotatebox[origin=c]{90}{\parbox{25mm}{\centering class-independent}}}
 & FPFH \cite{rusu2009fast} & 0.39 & 0.60 & 0.62 & 0.14 & 0.25 & 0.32 \\
 & Region growing & 0.59 & 0.70 & 0.71 & 0.19 & 0.34 & 0.38 \\
 & Rabbani et.al \cite{7710f6ecb9384791837478257854a571} & 0.62 & 0.71 & 0.72 & 0.17 & 0.31 & 0.36 \\
 & LRGNet \cite{lrgnet} & 0.67 & 0.74 & 0.75 & 0.25 & 0.41 & 0.43 \\
 & LRGNet+ local search \cite{lrgnet} & 0.68 & 0.75 & 0.76 & 0.34 & 0.44 &0.45 \\
 & \textbf{Region-Transformer} & \textbf{0.79} & \textbf{0.85} & \textbf{0.86} & \textbf{0.63} & \textbf{0.66} & \textbf{0.62} \\ [1ex] 
\hline 
\end{tabular}
\end{table*}

\begin{table*}[ht]
\caption{Comparison between models using S3DIS as training and Scannet as test.}\label{tab:example1} \centering 
\begin{tabular}{l c c c c c c c} 
\hline\hline 
\textbf{Type} & \textbf{Method} & \textbf{ARI} & \textbf{AMI} & \textbf{NMI} & \textbf{Precision} & \textbf{Recall} & \textbf{mIoU} \\ [0.5ex] 
\hline 
\multirow{5}{*}{\rotatebox[origin=c]{90}{\parbox{25mm}{\centering class-\\dependent}}}
 & PointNet \cite{qi2017pointnet} & 0.40 & 0.51 & 0.57 & 0.08 & 0.13 & 0.26 \\
 & PointNet++ \cite{pointnetplusplus} & 0.47 & 0.57 & 0.63 & 0.15 & 0.21 & 0.32 \\
 & 3D-BoNet \cite{yang2019learning} & 0.34 & 0.54 & 0.59 & 0.10 & 0.13 & 0.24\\
 & JSIS3D \cite{pham2019jsis3d} & 0.31 & 0.56 & 0.57 & 0.15 & 0.13 & 0.22 \\
 & Point Transformer \cite{pointtransformer} & 0.56 & 0.69 & 0.70 & \textbf{0.33} & 0.34 & 0.38 \\
\hline
\multirow{5}{*}{\rotatebox[origin=c]{90}{\parbox{25mm}{\centering class-independent}}}
 & FPFH \cite{rusu2009fast} & 0.28 & 0.51 & 0.53 & 0.10 & 0.14 & 0.26 \\
 & Region growing & 0.44 & 0.60 & 0.62 & 0.17 & 0.23 & 0.30 \\
 & Rabbani et.al \cite{7710f6ecb9384791837478257854a571} & 0.49 & 0.62 & 0.64 & 0.13 & 0.24 & 0.32 \\
 & LRGNet \cite{lrgnet} & 0.54 & 0.67 & 0.69 & 0.25 & 0.33 & 0.39 \\
 & LRGNet+ local search \cite{lrgnet} & 0.56 & 0.68 & 0.69 & 0.31 & 0.33 & 0.38 \\

 & \textbf{Region-Transformer} & \textbf{0.61} & \textbf{0.70} & \textbf{0.72} & 0.25 & \textbf{0.39} & \textbf{0.43} \\ [1ex] 
\hline 
\end{tabular}
\end{table*}

\begin{table*}[htb]
\caption{Computation Time analysis (seconds)}\label{tab:example1} \centering
\begin{tabular}{c c c c} 
\hline\hline 
Method & Minimum & Average & Maximum \\ [0.5ex] 
\hline 
Region growing & 0.4 & 4.8 & 18.6 \\ 
PointNet \cite{qi2017pointnet} & 0.1 & 0.6 & 2.5 \\
PointNet++ \cite{pointnetplusplus} & 0.1 & 0.9 & 3.5 \\
Rabbani et.al \cite{7710f6ecb9384791837478257854a571} & 0.3 & 4.6 & 18.3 \\
3D-BoNet \cite{yang2019learning} & 1.5 & 14.1 & 69.3 \\
FPFH \cite{rusu2009fast} & 0.5 & 4.6 & 17.8 \\
LRGNet \cite{lrgnet} & 0.8 & 64.9 & 620.9 \\
JSIS3D \cite{pham2019jsis3d} & 1.0 & 539.2 & 16713.9 \\
\textbf{Region-Transformer} & 1.5 & 57.4 & 311.5 \\ [1ex] 
\hline 
\end{tabular}
\end{table*}

The study conducted extensive experiments to evaluate the Region-Transformer, comparing it against previous segmentation methods using the S3DIS and ScanNet datasets. These datasets represent different environments: S3DIS focuses on office settings, while ScanNet covers home environments. This diverse testing revealed that the Region-Transformer excelled over other methods across almost every metric, as shown in Tables 1 and 2.

A significant aspect of this success is attributed to the model's use of local neighborhood information combined with a self-attention mechanism. This approach was particularly effective compared to local search techniques like those in LRGNet \cite{lrgnet}. The research underscores the advantage of applying the self-attention mechanism in a region-based, class-agnostic approach for point cloud segmentation. Unlike methods trained with semantic label information that showed diminished performance when applied to a different dataset, the Region-Transformer demonstrated robust generalization capabilities.

Regarding specific evaluation metrics, the Normalized Mutual Information(NMI) metric assesses the similarity between two clusters, with a value range of 0 to 1. A high NMI score for the Region-Transformer indicates a reduction in the entropy of instance labels and an improvement in the under-segmentation of instance labels. The method's high NMI is attributed to its ability to predict pure clusters that closely match the ground truth. Similar to NMI but adjusted for random chance, AMI ranges from -1 to 1. The Region-Transformer scored high on AMI, signifying its efficiency in creating pure clusters and solving over-segmentation problems. AMI accounts for the number of clusters and dataset size, discounting chance normalization. From -1 to 1, ARI is related to accuracy in measuring the percentage of correct predictions. It corrects for a change from the rand index and is particularly useful for considering unbalanced clustering.

The paper emphasizes that while NMI and AMI are effective for evaluating clustering purity and similarity, they have limitations. For instance, NMI can increase with the number of clusters regardless of actual mutual information. Similarly, AMI might be biased towards unbalanced clustering solutions. Hence, including ARI as a metric provides a more comprehensive and balanced evaluation.

In the qualitative evaluation, the Region-Transformer demonstrates marked improvements in segmenting indoor scenes, as shown in Figures 3 and 4. It effectively resolves under-segmentation on smooth surfaces like floors and accurately distinguishes objects of varying shapes and sizes. However, challenges arise in wall segmentation due to uneven surfaces and corner over-segmentation. Despite these issues, its performance in differentiating unique instances in environments like the S3DIS indoor scenes is notable.

The model's adaptability to large-scale scenes is a key strength. Initially trained on homes and offices, the Region-Transformer shows remarkable capability in segmenting larger environments, including factories and large buildings, as illustrated in Figure 5. This is essential for applications in self-driving cars and digital twinning, demonstrating its practical utility in handling complex, large-scale environments without prior object knowledge in general.

Furthermore, the Region-Transformer significantly improves computational efficiency, particularly in inference time, compared to other class-agnostic, region-based segmentation approaches. Despite the iterative nature of its process, it maintains better efficiency, a finding supported by the average inference time analysis of 50 S3DIS datasets presented in Table 3. This balance of accuracy and computational speed makes it suitable for real-time applications, highlighting its potential in accuracy and efficiency.

\section{\uppercase{Conclusions}}
\label{sec:conclusions}
We propose a novel region-based transformer model called Region-Transformer for performing class-agnostic point cloud segmentation. Experiments demonstrate that combining self-attention with an iterative region-growing approach significantly improves segmentation performance. Specifically, attention mechanisms effectively capture local contextual relationships between points missed by previous region growth methods.
Key advantages of the proposed approach include:
\begin{itemize}
    \item Attention on local point neighborhoods enables capturing finer feature relationships versus global context. This aids in precisely determining segmentation boundaries.
    \item The region growth formulation allows flexible, adaptive segmentation based on progressively learned point neighborhood relationships, avoiding strong assumptions.
    \item The method avoids dependence on semantic class labels, enabling new object segmentation.
\end{itemize}

The promising performance and flexibility of Region-Transformer represent an important step toward practical point cloud segmentation without prior knowledge. Potential real-world applications span robotic perception, autonomous navigation, digital twinning, and augmented reality.

Future avenues for improving Region-Transformer include reducing training and inference times via model compression techniques tailored for transformers. New spatial attention operators could also be designed to capture geometric relationships in point clouds. An exciting research direction involves extending the approach to perform video segmentation on dynamic point cloud sequences containing moving objects.

\section*{\uppercase{Acknowledgements}}

This project is supported in part by NSF grant OIA-1946231 and NASA. We are grateful to Mr. Marc Aubanel for his feedback and providing data for this research.


\bibliographystyle{apalike}
{\small
\bibliography{example}}

\begin{thebibliography}{}

\bibitem[Ahn et~al., 2022]{9687584}
Ahn, P., Yang, J., Yi, E., Lee, C., and Kim, J. (2022).
\newblock Projection-based point convolution for efficient point cloud segmentation.
\newblock {\em IEEE Access}, 10:15348–15358.

\bibitem[Armeni et~al., 2017]{s3dis}
Armeni, I., Sax, A., Zamir, A.~R., and Savarese, S. (2017).
\newblock Joint 2d-3d-semantic data for indoor scene understanding.
\newblock {\em ArXiv e-prints}.

\bibitem[Asao and Ike, 2022]{asao2022curvature}
Asao, Y. and Ike, Y. (2022).
\newblock Curvature of point clouds through principal component analysis.

\bibitem[Chen et~al., 2021a]{lrgnet}
Chen, J., Kira, Z., and Cho, Y.~K. (2021a).
\newblock Lrgnet: Learnable region growing for class-agnostic point cloud segmentation.
\newblock {\em IEEE Robotics and Automation Letters}, 6(2):2799--2806.

\bibitem[Chen et~al., 2021b]{autonomous}
Chen, S., Liu, B., Feng, C., Vallespi-Gonzalez, C., and Wellington, C. (2021b).
\newblock 3d point cloud processing and learning for autonomous driving: Impacting map creation, localization, and perception.
\newblock {\em IEEE Signal Processing Magazine}, 38(1):68--86.

\bibitem[Dai et~al., 2017]{scannet}
Dai, A., Chang, A.~X., Savva, M., Halber, M., Funkhouser, T., and Nießner, M. (2017).
\newblock Scannet: Richly-annotated 3d reconstructions of indoor scenes.
\newblock In {\em Proc. Computer Vision and Pattern Recognition (CVPR)}. IEEE.

\bibitem[Dimitrov and Golparvar-Fard, 2015]{81fc0f1e9d33404e8279328c15c08bca}
Dimitrov, A. and Golparvar-Fard, M. (2015).
\newblock Segmentation of building point cloud models including detailed architectural/structural features and mep systems.
\newblock {\em Automation in Construction}, 51(C):32--45.
\newblock Publisher Copyright: © 2014 Elsevier B.V. All rights reserved.

\bibitem[Engelmann et~al., 2019]{Engelmann_2019}
Engelmann, F., Kontogianni, T., Schult, J., and Leibe, B. (2019).
\newblock {\em Know what your neighbors do: 3d semantic segmentation of point clouds}, page 395–409.
\newblock Springer International Publishing.

\bibitem[Gumhold et~al., 2001]{Gumhold2001FeatureEF}
Gumhold, S., Wang, X., and Macleod, R. (2001).
\newblock Feature extraction from point clouds.
\newblock In {\em International Meshing Roundtable Conference}.

\bibitem[Guo et~al., 2015]{guoyulan}
Guo, Y., Bennamoun, M., Sohel, F., Lu, M., Wan, J., and Kwok, N. (2015).
\newblock A comprehensive performance evaluation of 3d local feature descriptors.
\newblock {\em International Journal of Computer Vision}, 116.

\bibitem[Guo et~al., 2020]{guo2020deep}
Guo, Y., Wang, H., Hu, Q., Liu, H., Liu, L., and Bennamoun, M. (2020).
\newblock Deep learning for 3d point clouds: A survey.

\bibitem[Gyawali, 2023]{gyawali2023lrtransformer}
Gyawali, D. (2023).
\newblock Lrtransformer: Learn-region transformer for object-agnostic point cloud segmentation.
\newblock Master's thesis, Louisiana State University.

\bibitem[Hu et~al., 2019]{hu2019learning}
Hu, P., Held, D., and Ramanan, D. (2019).
\newblock Learning to optimally segment point clouds.

\bibitem[Kang et~al., 2020]{isprs-archives-XLII-3-W10-153-2020}
Kang, C.~L., Wang, F., Zong, M.~M., Cheng, Y., and Lu, T.~N. (2020).
\newblock Research on improved region growing point cloud algorithm.
\newblock {\em The International Archives of the Photogrammetry, Remote Sensing and Spatial Information Sciences}, XLII-3/W10:153--157.

\bibitem[Ling et~al., 2021]{robotics}
Ling, C.~F., Dang, S.~W., Zhang, C.~Y., and Chen, Y. (2021).
\newblock Research and application of semantic point cloud on indoor robots.
\newblock In {\em 2021 5th International Conference on Communication and Information Systems (ICCIS)}, pages 108--113.

\bibitem[Mirzaei et~al., 2022]{digitaltwin}
Mirzaei, K., Arashpour, M., Asadi, E., et~al. (2022).
\newblock Automatic generation of structural geometric digital twins from point clouds.
\newblock {\em Sci Rep}, 12:22321.

\bibitem[Nguyen and Le, 2013]{6758588}
Nguyen, A. and Le, B. (2013).
\newblock 3d point cloud segmentation: A survey.
\newblock In {\em 2013 6th IEEE Conference on Robotics, Automation and Mechatronics (RAM)}, page 225–230.

\bibitem[Nunes et~al., 2022]{9813403}
Nunes, L., Chen, X., Marcuzzi, R., Osep, A., Leal-Taixe, L., and Stachniss, C. (2022).
\newblock 3d point cloud clustering with learnable robust geometric constraints.

\bibitem[Pham et~al., 2019]{pham2019jsis3d}
Pham, Q.-H., Nguyen, D.~T., Hua, B.-S., Roig, G., and Yeung, S.-K. (2019).
\newblock Jsis3d: Joint semantic-instance segmentation of 3d point clouds with multi-task pointwise networks and multi-value conditional random fields.

\bibitem[Placitelli and Gallo, 2011]{ar}
Placitelli, A.~P. and Gallo, L. (2011).
\newblock Low-cost augmented reality systems via 3d point cloud sensors.
\newblock In {\em 2011 Seventh International Conference on Signal Image Technology \& Internet-Based Systems}, pages 188--192.

\bibitem[Qi et~al., 2017a]{qi2017pointnet}
Qi, C.~R., Su, H., Mo, K., and Guibas, L.~J. (2017a).
\newblock Pointnet: Deep learning on point sets for 3d classification and segmentation.

\bibitem[Qi et~al., 2017b]{pointnetplusplus}
Qi, C.~R., Yi, L., Su, H., and Guibas, L.~J. (2017b).
\newblock Pointnet++: Deep hierarchical feature learning on point sets in a metric space.

\bibitem[Rabbani et~al., 2006]{7710f6ecb9384791837478257854a571}
Rabbani, T., van~den Heuvel, F., and Vosselman, G. (2006).
\newblock Segmentation of point clouds using smoothness constraints.
\newblock In Maas, H. and Schneider, D., editors, {\em ISPRS 2006 : Proceedings of the ISPRS commission V symposium}, volume~35, pages 248--253. International Society for Photogrammetry and Remote Sensing (ISPRS).
\newblock ISPRS commission V symposium : image.

\bibitem[Rusu et~al., 2009]{rusu2009fast}
Rusu, R.~B., Blodow, N., and Beetz, M. (2009).
\newblock Fast point feature histograms (fpfh) for 3d registration.
\newblock In {\em 2009 IEEE International Conference on Robotics and Automation}, pages 3212--3217.

\bibitem[Sharma et~al., 2020]{sharma2020classagnostic}
Sharma, A., Khan, N., Sundaramoorthi, G., and Torr, P. (2020).
\newblock Class-agnostic segmentation loss and its application to salient object detection and segmentation.

\bibitem[Vaswani et~al., 2023]{vaswani2023attention}
Vaswani, A., Shazeer, N., Parmar, N., Uszkoreit, J., Jones, L., Gomez, A.~N., Kaiser, L., and Polosukhin, I. (2023).
\newblock Attention is all you need.

\bibitem[Wang and Yuan, 2010]{5703032}
Wang, L. and Yuan, B. (2010).
\newblock Curvature and density based feature point detection for point cloud data.
\newblock In {\em IET 3rd International Conference on Wireless, Mobile and Multimedia Networks (ICWMNN 2010)}, page 377–380.

\bibitem[Xiao et~al., 2023]{Xiao_2023}
Xiao, A., Huang, J., Guan, D., Zhang, X., Lu, S., and Shao, L. (2023).
\newblock Unsupervised point cloud representation learning with deep neural networks: A survey.
\newblock {\em IEEE Transactions on Pattern Analysis and Machine Intelligence}, pages 1--20.

\bibitem[Xie et~al., 2021]{xie2021unseen}
Xie, C., Xiang, Y., Mousavian, A., and Fox, D. (2021).
\newblock Unseen object instance segmentation for robotic environments.

\bibitem[Yang et~al., 2019]{yang2019learning}
Yang, B., Wang, J., Clark, R., Hu, Q., Wang, S., Markham, A., and Trigoni, N. (2019).
\newblock Learning object bounding boxes for 3d instance segmentation on point clouds.

\bibitem[Zhao et~al., 2020]{zhao2020exploring}
Zhao, H., Jia, J., and Koltun, V. (2020).
\newblock Exploring self-attention for image recognition.

\bibitem[Zhao et~al., 2021a]{pointtransformer}
Zhao, H., Jiang, L., Jia, J., Torr, P., and Koltun, V. (2021a).
\newblock Point transformer.

\bibitem[Zhao et~al., 2021b]{zhao2021fewshot}
Zhao, N., Chua, T.-S., and Lee, G.~H. (2021b).
\newblock Few-shot 3d point cloud semantic segmentation.

\bibitem[Zhao et~al., 2019]{s19051248}
Zhao, R., Pang, M., Liu, C., and Zhang, Y. (2019).
\newblock Robust normal estimation for 3d lidar point clouds in urban environments.
\newblock {\em Sensors}, 19(5).

\bibitem[Zhou et~al., 2021]{zhou2021fast}
Zhou, J., Jin, W., Wang, M., Liu, X., Li, Z., and Liu, Z. (2021).
\newblock Fast and accurate normal estimation for point cloud via patch stitching.

\end{thebibliography}

\end{document}